# The Illusion of Clinical Reasoning: A Benchmark Reveals the Pervasive Gap in Vision-Language Models for Clinical Competency

Dingyu Wang[1,2,3#], Zimu Yuan[1,2,3#], Jiajun Liu[1,2,3], Shanggui Liu[1,2,3], Nan Zhou[4,5], Tianxing Xu[4,5], Di Huang[4,5*], Dong Jiang[1,2,3*]

1. Department of Sports Medicine, Peking University Third Hospital, Institute of Sports Medicine of Peking University, Beijing, China.
2. Beijing Key Laboratory of Sports Injuries, Beijing, China.
3. Engineering Research Center of Sports Trauma Treatment Technology and Devices, Ministry of Education, Beijing, China.
4. State Key Laboratory of Complex and Critical Software Environment, Beihang University, Beijing, China
5. School of Computer Science and Engineering, Beihang University, Beijing, China

# These authors contributed equally to this study.

*These are corresponding authors.

## Abstract

Background: The rapid integration of foundation models into clinical practice and public health necessitates a rigorous evaluation of their true clinical reasoning capabilities beyond narrow examination success. Current benchmarks, typically based on medical licensing exams or curated vignettes, fail to capture the integrated, multimodal reasoning essential for real-world patient care.

Methods: We developed the Bones and Joints (B&J) Benchmark, a comprehensive evaluation framework comprising 1,245 questions derived from real-world patient cases in orthopedics and sports medicine. This benchmark assesses models across 7 tasks that mirror the clinical reasoning pathway, including knowledge recall, text and image interpretation, diagnosis generation, treatment planning, and rationale provision. We evaluated eleven vision-language models (VLMs) and six large language models (LLMs), comparing their performance against expert-derived ground truth.

Results: Our results demonstrate a pronounced performance gap between task types. While state-of-the-art models achieved high accuracy, exceeding 90%, on structured multiple-choice questions, their performance markedly declined on open-ended tasks requiring multimodal integration, with accuracy scarcely reaching 60%. VLMs demonstrated substantial limitations in interpreting medical images and frequently exhibited severe text-driven hallucinations, often ignoring contradictory visual evidence. Notably, models specifically fine-tuned for medical applications showed no consistent advantage over general-purpose counterparts.

Conclusions: Current artificial intelligence models are not yet clinically competent for complex, multimodal reasoning. Their safe deployment should currently be limited to supportive, text-based roles. Future advancement in core clinical tasks awaits fundamental breakthroughs in multimodal integration and visual understanding.

# Introduction

The emergence of next-generation foundation models is reshaping the field of medical artificial intelligence (AI). Two recent technological advances, the development of large reasoning models and the maturation of vision-language models (VLMs), have enabled AI systems to address complex medical tasks that require sophisticated planning, integration of multimodal information, and other high-level cognitive skills. State-of-the-art (SOTA) models have demonstrated performance comparable to or surpassing human experts on multiple medical benchmarks[1, 2]. Consequently, these foundation models are being rapidly integrated into clinical workflows, where they are tasked with summarizing records, generating diagnostic reports, and providing decision support[3, 4]. Moreover, these tools are now widely available to the public, who are turning to them for everyday health-related questions and initial symptom assessments. Patients can readily input their symptoms or medical history into AI-powered chatbots, which offer 24/7 access to immediate medical guidance[5, 6].

While developers and social media often highlight the remarkable achievements of AI models and their perceived superiority over clinicians, a crucial limitation tends to be overlooked: these impressive results are largely derived from medical licensing examinations, narrow question-answering datasets, curated and constrained clinical vignettes[7-9], which do not reflect the integrated and nuanced nature of real-world clinical reasoning[10, 11]. It must be acknowledged that passing such medical examination is merely the first step toward becoming a clinician. Beyond the acquisition of knowledge, a clinician must synthesize information from diverse sources, including clinical notes, physical examinations, laboratory results, and medical images, and apply evidence-based reasoning within diagnostic pathways. This gap raises a critical question: can contemporary AI models truly achieve clinical competence, especially when faced with multimodal data and conflicting information in real healthcare environments? In the absence of robust evaluation methods for such capabilities, the promise of medical AI remains built on unstable foundations, potentially jeopardizing patient safety, the very objective it aims to enhance[12, 13]. This risk is particularly acute for the non-professional public, who may lack the expertise to recognize model inaccuracies or biases.

To close this critical gap, the development of a more rigorous evaluation framework is imperative. Such a benchmark must be built upon clinical cases that reflect the true complexities of medicine. Ideal test cases should possess a broad disease spectrum, including both common and rare conditions, to thoroughly assess model generalization. They must be inherently multimodal, relying on textual and visual evidence as equally critical components, thereby compelling genuine information integration. Furthermore, the clinical reasoning pathway should be of intermediate complexity, challenging a model's capacity for sequential planning and evidence weighing, while the optimal management should involve personalized decision-making that adapts to individual patient factors beyond standardized protocols.

Clinical cases in orthopedics and sports medicine serve as a paradigmatic example that fulfills these criteria. This domain addresses a massive global burden, affecting over 1.71 billion people and representing a leading cause of pain and disability[14-17]. The diagnostic process is fundamentally multimodal, demanding the synthesis of patient history, physical examination findings, and radiological images. Moreover, treatment planning is highly personalized, tailored to individual patient demands and activity levels[18]. This combination of mandatory multimodal fusion and a complex reasoning pathway makes musculoskeletal disorders a uniquely demanding and revealing testbed for models.

Here, we present the Bones and Joints (B&J) Benchmark, a comprehensive framework designed to probe the authentic clinical reasoning abilities of modern VLMs and large language models (LLMs). Comprising 1,245 questions derived primarily from unpublished real-world patient cases from eight tertiary hospitals, our benchmark structures evaluation along the clinical reasoning pathway through seven interconnected tasks. These range from basic knowledge recall and unimodal interpretation to advanced multimodal challenges like diagnosis generation, treatment planning, and providing the underlying diagnostic and therapeutic rationale (Fig. 1). We evaluated 11 leading VLMs and six LLMs using this benchmark, comparing their outputs against expert-provided final decisions and chain-of-thought reasoning.

Our results reveal a pronounced capability gap. While SOTA models achieved impressive accuracy exceeding 90% on structured multiple-choice questions, their performance markedly collapsed when confronted with multimodal open-ended tasks requiring diagnostic reasoning or treatment planning, where accuracy rates scarcely reached 60%. Furthermore, VLMs demonstrated substantial limitations in medical image interpretation and frequently exhibited severe text-driven hallucinations when processing multimodal clinical data. Notably, models designed specifically for medical applications showed no consistent advantage over their general-purpose counterparts. These findings clarify the current limits of AI in clinical reasoning and underscore the need for fundamental advances before safe deployment in core clinical tasks.

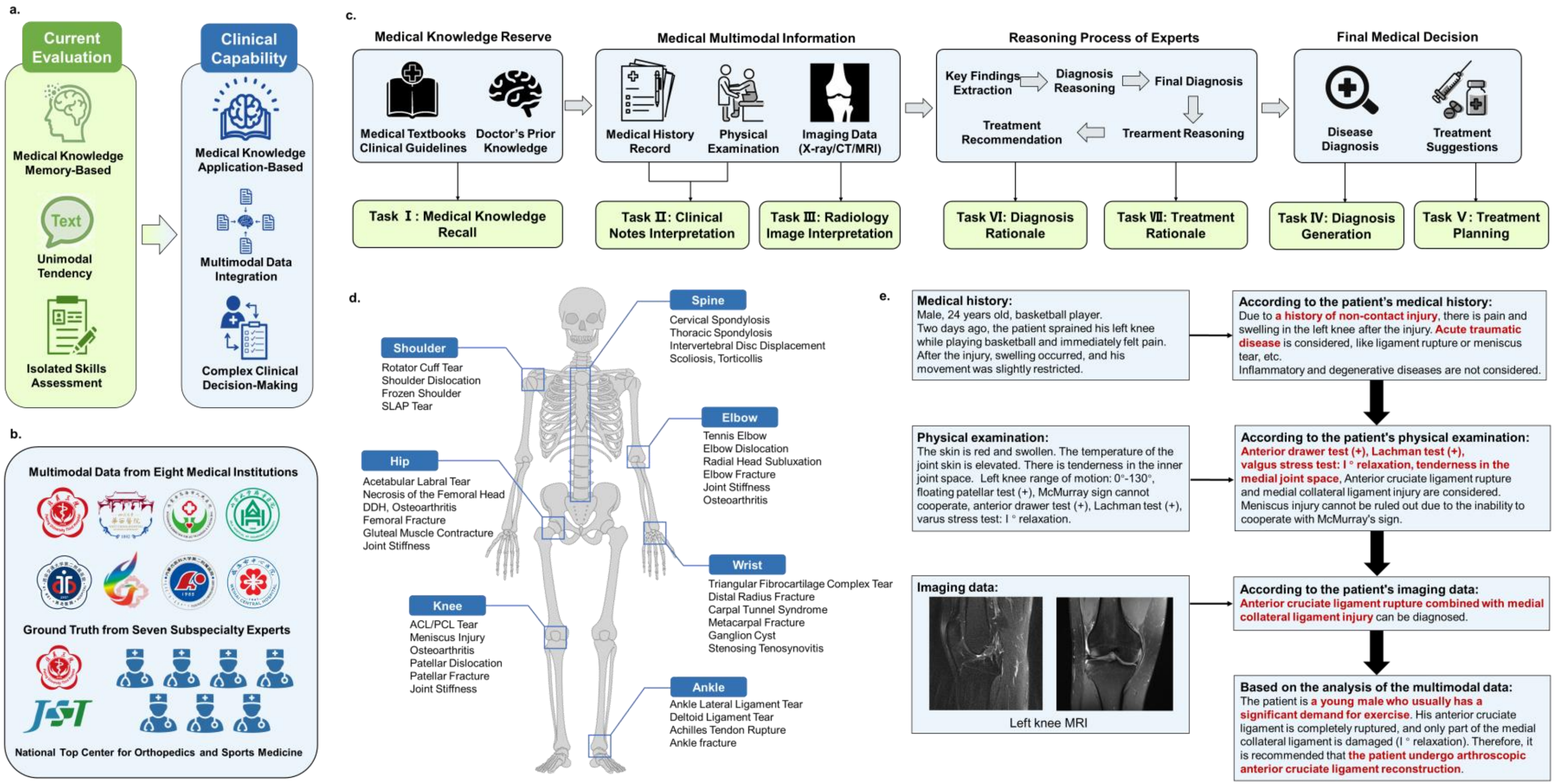

Figure 1. Overview of our study. **a)** The current gap between AI evaluation and the real clinical capability of clinicians. **b)** The benchmark data sources: all cases were sourced from eight hospitals, and the answers to the questions were compiled by seven professional orthopedic and sports medicine experts as the ground truth. **c)** Design idea for the benchmark: along the complete cognitive process of clinicians in clinical practice, the benchmark has designed seven types of questions to separately evaluate model's competency. **d)** Comprehensive Anatomical and Disease Coverage: The benchmark covers 7 orthopedic anatomical sites, and 42 common orthopedic and sports medicine diseases. **e)** An example of reasoning process by a clinician from obtaining multimodal information to outputting the final disease diagnosis and treatment suggestions. The reasoning ability of the model was evaluated according to ground truth in such format.

# Methods

## Data source

The B&J Benchmark was prospectively constructed from two distinct data sources to ensure a comprehensive evaluation of both foundational knowledge and applied clinical reasoning.

**Clinician Certification Exams.** To assess core medical knowledge, we utilized official question banks from two national certification exams required for orthopedic specialists in China: The National Medical Licensing Examination (NMLEC) and the Orthopedic Attending Physician Qualification Examination. We selected multiple-choice questions specifically related to orthopedics from these sources, forming the basis for tasks evaluating knowledge recall and comprehension.

**Real Clinical Cases.** To evaluate clinical reasoning, we collected real-world cases from eight medical centers across different regions. All cases were collected prospectively to ensure uniqueness and prevent data contamination from existing public datasets.

Our goal was to include cases with complete clinical information: encompassing medical history, physical examination, laboratory tests, and imaging data (X-rays, CT, MRI), covering a broad spectrum of musculoskeletal diseases across seven anatomical regions (knee, hip, ankle, elbow, shoulder, wrist, and spine). The cases were screened using the following exclusion criteria: 1) lack of essential imaging data required for diagnosis; 2) high textual similarity with other cases in the cohort; 3) pre-existing explicit diagnostic conclusions in the medical records, which would hinder the effective evaluation of the model's true diagnostic reasoning ability.

Following this screening, 203 high-quality cases covering 42 common musculoskeletal diseases were retained. These cases subsequently underwent a standardized processing protocol: 1) De-identification: all personally identifiable information (e.g., patient names, record numbers) was removed to protect privacy; 2) Structuring: unstructured clinical narratives (e.g., from medical histories) were extracted and organized into standardized fields to ensure homogeneous model input;

3) Expert Review: A team of senior orthopedic experts manually reviewed all processed data to correct errors and guarantee accuracy.

**Development of Benchmark**

The B&J Benchmark was designed to comprehensively mirror the entire clinical workflow, structuring its evaluation around seven core tasks that reflect a clinician's reasoning process. These tasks progress from unimodal knowledge and interpretation to complex multimodal reasoning.

The benchmark begins with foundational, unimodal tasks:

- ✓ Task I (Knowledge Recall), derived from national examination question banks, includes 129 multiple-choice questions assessing core concept and theory knowledge about diagnosis and treatment.
- ✓ Task II (Clinical Note Interpretation), also from national examinations, comprises 101 multiple-choice questions that require predicting a diagnosis or treatment from a summarized clinical note.
- ✓ Task III (Radiological Image Interpretation), sourced from real clinical cases, consists of 203 multiple-choice questions where the model must diagnose a condition directly from an X-ray, MRI, or CT image.

The evaluation then advances to open-ended, multimodal reasoning tasks, all derived from real clinical cases:

- ✓ Tasks IV & V (Diagnosis Generation and Reasoning) require the model to generate a complete diagnosis and provide the underlying chain-of-thought by integrating full multimodal information.
- ✓ Tasks VI & VII (Treatment Planning and Reasoning) challenge the model to recommend a therapy plan and output its clinical rationale.

To accommodate the text-only input of LLMs, all imaging data in the open-ended questions were replaced with their corresponding radiology reports, while all other question content remained unchanged.

Finally, an additional adversarial task was designed to test multimodal integrity:

- ✓ Task VIII (Identify Text-Image Inconsistencies) presents deliberately mismatched clinical text and medical images (e.g., a knee injury history paired with a shoulder

MRI). The model is prompted to generate a diagnosis and rationale, testing its ability to recognize cross-modal contradictions rather than defaulting to text-driven hallucinations.

## Ground Truth

We established a comprehensive Ground Truth generation process, which is led by seven senior experts with over ten years of clinical experience. These experts are from Peking University Third Hospital and Beijing Jishuitan Hospital, both of which are leading institutions in the field of orthopedics and sports medicine in China, renowned for their academic and clinical excellence. The composition of the expert team members covers multiple subspecialties within orthopedics and sports medicine, including shoulder and elbow surgery, foot and ankle surgery, spine surgery, and hand surgery, etc. Each expert has focused on 2-3 specific anatomical regions, thereby accumulating rich diagnostic and therapeutic experience in their respective areas of expertise.

During the annotation process, we designed a rigorous review mechanism for different question types. For multiple-choice questions, the expert team reviewed and revised the reference answers from the question banks in accordance with the latest clinical practice guidelines and expert consensus. For open-ended questions, the experts annotated the entire clinical reasoning process for each case. This involved a detailed analysis of the patient's medical history, physical examination and imaging data to construct a clear diagnostic logic chain and provide a precise diagnosis. Combined with the individual characteristics of the patients and the experts' rich clinical experience, they then proposed personalized treatment plans. To ensure the reliability and consistency of the annotated results, a dual-expert review process was implemented for all case annotations. One expert performed the initial annotation, and another expert conducted an independent review. In cases of disagreement, a consensus was reached through full discussion to determine the final answer.

## Model Selection

To conduct a comprehensive and representative evaluation, we selected 17 AI models, covering four categories: General-purpose VLMs, Medical-specific VLMs, General-purpose LLMs, and Medical-specific LLMs. We give priority to open-source and closed-source models that adopt mainstream architectures (such as Transformer or MoE) and perform outstandingly in current medical benchmarks (such as MIMIC-CXR, MedQA). Detailed information about the selected models can be found in Table 1.

Table 1: Detailed information of Selected Models for Evaluation.

| Model | Size | Input modalities | Output modalities | Open-Source | Link | Reference |
|---|---|---|---|---|---|---|
| GLM-4V-9B | 9B | Text + Image | Text | Yes | https://huggingface.co/THUDM/glm-4v-9b | https://arxiv.org/abs/2406.12793 |
| Qwen2-VL | 7B | Text + Image | Text | Yes | https://huggingface.co/Qwen/Qwen2-VL-7B-Instruct | https://arxiv.org/abs/2409.12191 |
| MiniCPM-V2.6 | 8B | Text + Image | Text | Yes | https://huggingface.co/openbmb/MiniCPM-V-2_6 | https://arxiv.org/abs/2408.01800 |
| Llama-3.2-Vision | 11B | Text + Image | Text | Yes | https://huggingface.co/meta-llama/Llama-3.2-11B-Vision | https://arxiv.org/abs/2204.05149 |
| GPT-4o | - | Text + Image | Text | No | https://chatgpt.com/ | https://arxiv.org/abs/2410.21276 |
| Claude3.5-Sonnet | - | Text + Image | Text | No | https://www.anthropic.com/claude/sonnet | https://assets.anthropic.com/m/1cd9d098ac3e6467/original/Claude-3-Model-Card-October-Addendum.pdf |
| DeepSeek-VL2 | 4.5B | Text + Image | Text | Yes | https://github.com/deepseek-ai/DeepSeek-VL2 | https://arxiv.org/abs/2412.10302 |
| Med-Flamingo | 9B | Text + Image | Text | Yes | https://huggingface.co/med-flamingo/med-flamingo | https://arxiv.org/abs/2307.15189 |
| LLaVA-Med | 7B | Text + Image | Text | Yes | https://huggingface.co/microsoft/llava-med-v1.5-mistral-7b | https://arxiv.org/abs/2306.00890 |
| MedVInT | - | Text + Image | Text | Yes | https://github.com/xiaoman-zhang/PMC-VQA/tree/master/src/MedVInT_TD | https://arxiv.org/abs/2305.10415 |
| MiniGPT-Med | - | Text + Image | Text | Yes | https://github.com/Vision- | https://arxiv.org/abs/2407.04106 |
| | | | | | CAIR/MiniGPT-Med | |
| DeepSeek-R1-0528 | - | Text | Text | Yes | https://github.com/deepseek-ai/DeepSeek-R1 | https://arxiv.org/abs/2501.12948 |
| Qwen2.5-32B | 32B | Text | Text | Yes | https://huggingface.co/Qwen/Qwen2.5-32B | https://arxiv.org/abs/2407.10671 |
| GLM-4-9B | 9B | Text | Text | Yes | https://huggingface.co/THUDM/glm-4-9b | https://arxiv.org/abs/2406.12793 |
| MedGPT | 13B | Text | Text | Yes | https://huggingface.co/Medlinker/Medgpt | https://www.medlinker.com/index/medgpt.html |
| MedFound | 8B | Text | Text | Yes | https://github.com/medfound/medfound | https://www.nature.com/articles/s41591-024-03416-6 |
| Baichuan-M2 | 32B | Text | Text | Yes | https://github.com/baichuan-inc/Baichuan-M2-32B | https://arxiv.org/abs/2509.02208 |

**Model deployment**

We have adopted different deployment strategies for open-source and closed-source models. The open-source model is deployed through the Hugging Face Transformers library, and its source code and weights are downloaded from official sources. Closed-source models are invoked through the API interface officially provided by them. All experiments were conducted using NVIDIA A100 GPUs to ensure high computational throughput and memory efficiency.

**Evaluation Protocol**

All models were evaluated under a strict zero-shot setting. This means they were applied directly to downstream tasks without any task-specific fine-tuning, in-context examples, or parameter updates. This methodology directly mirrors a real-world clinical or public application, where a model must perform reliably without manual calibration.

We designed a unified prompt (see in Supplementary Appendix) for each task consistently across all models, which eliminates prompt engineering as a confounding variable, ensuring that performance differences observed can be attributed directly to the models' abilities rather than variations in input instructions. This evaluation framework, encompassing 17 models, can analyze the performance differences among models of different architectures and between models in single-modal and multi-modal information processing in complex clinical tasks.

**Evaluation Metrics**

To objectively evaluate the overall performance of each model in the field of orthopedic and sport medicine specialties, we have designed a multi-level evaluation framework, adopting different evaluation methods for different question types (Fig. 2).

For MCQs in the dataset, we adopt Accuracy as the evaluation metric.

For the evaluation of open-ended questions, we have designed a dual evaluation process that combines automated and human evaluation.

**Automated Evaluation.** We utilize a powerful LLM (GPT-4o) as the referee for automated evaluation. Previous studies have shown that using GPT as a judge can exhibit consistency close to that of human evaluation, making it particularly suitable for this task[19, 20]. The evaluation process is as follows: Input the answers generated by the model to be evaluated and our reference answers (ground truth) to GPT-4o simultaneously. The GPT-4o is required to determine whether the core semantics of the answers generated by the model are consistent with the reference answers, and output one of the three categories of determination results: Correct, Incorrect, or Unclear. The prompt for this evaluation is attached to in Supplementary Appendix. To reduce the element of chance, the assessment of each question is independently repeated three times. The average of the three results is taken as the final score. Ultimately, the total score of the model is presented in the form of accuracy.

**Manual evaluation.** We randomly sampled 120 items from the short-answer questions, including 30 questions each for Task IV (Diagnoses Generation), Task V (Diagnostic reasoning), Task VI (Treatment planning) and Task VII (Treatment reasoning). The evaluation was performed independently by two senior orthopedic physicians, each with over ten years of clinical experience and no involvement in the benchmark's construction. A double-blind assessment was implemented, where the identity of the model generating each response was concealed from the reviewers. Each response was assessed using a binary judgment of "Correct" or "Incorrect". Interrater reliability was calculated using Cohen's Kappa coefficient to quantify the agreement of the judgements among experts. In the randomly selected human evaluation dataset, the Kappa coefficient between the two senior experts was 0.95, indicating that the experts had reached a complete consensus. For all questions with inconsistent evaluations, the final decision will be made after consultation and discussion chaired by a third expert.

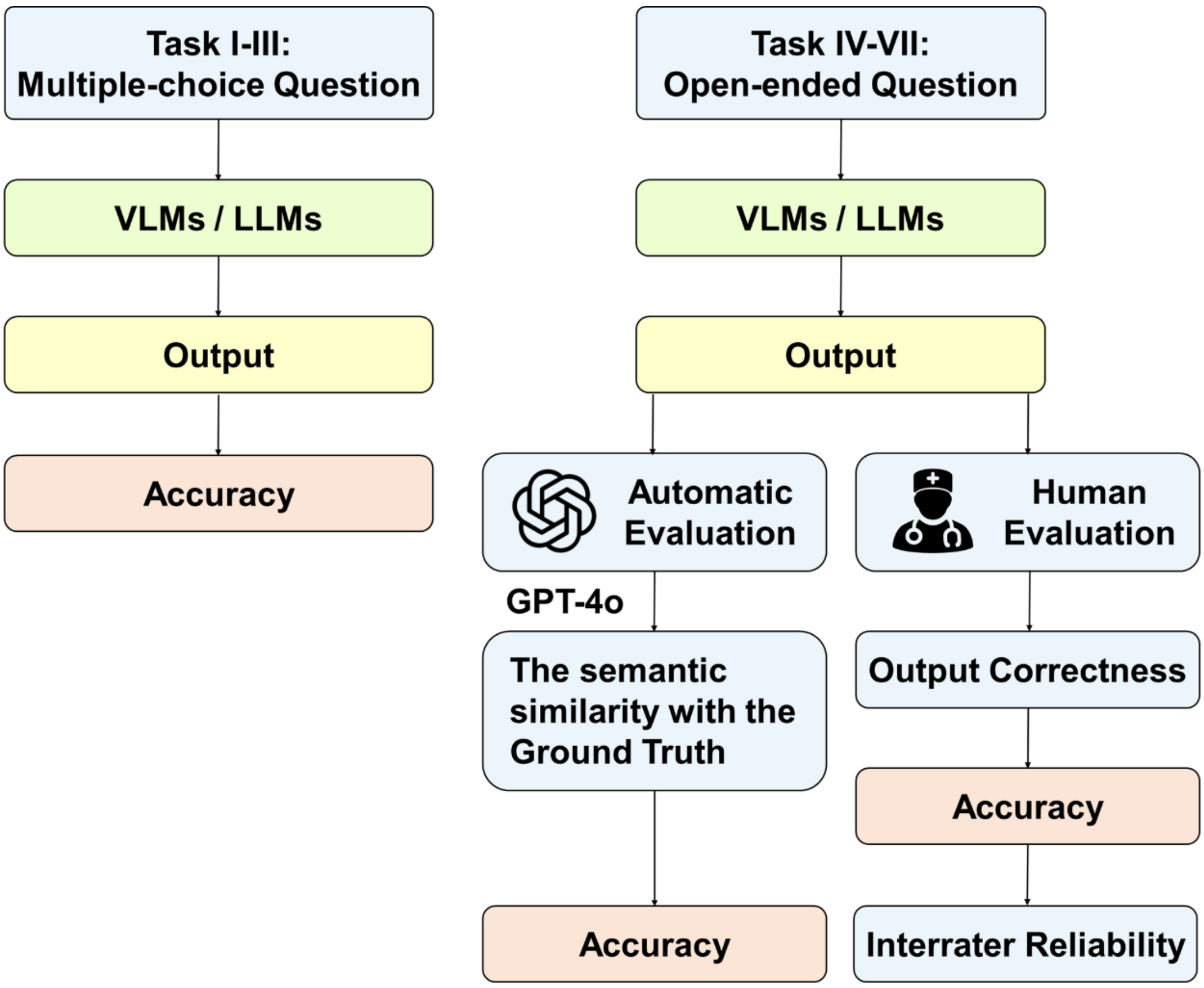

Figure 2. The Framework of Model Evaluation.

For Task VIII (Text-Image Inconsistency Identification), a response was considered successful only if the model identified the contradiction between the radiological text and the corresponding image. A failure was recorded if the model ignored the visual evidence and generated a diagnosis based solely on the erroneous text prompt. A canonical example of this failure mode is a model describing an "anterior cruciate ligament tear" when presented with a shoulder MRI, demonstrating an over-reliance on textual input without grounding its response in the provided image.

**Statistical Analyses**

All statistical analyses were conducted using GraphPad Prism software (Version 9.0.0, San Diego, California USA). The accuracy of each model across all question types was calculated, accompanied by 95% confidence intervals (CIs). These computations were performed in Prism 9, utilizing the number of correct and incorrect responses previously tabulated in Microsoft Excel.

To evaluate the statistical significance of differences in accuracy among the different models, a non-parametric analytical approach was employed. Initially, a Kruskal-Wallis test was conducted to assess the overall difference among the models. Following this, pairwise comparisons between models were performed using Dunn's multiple comparisons test as a post-hoc analysis. Adjusted P-values were obtained from Dunn's test to account for multiple comparisons. A P-value of less than 0.05 was considered to be statistically significant.

**Data availability**

The dataset including multiple-choice questions and open-ended questions, and their ground truth labels have been made available through HuggingFace library (https://huggingface.co/datasets/PUTH2025/Bones_and_Joints_Benchmark).

# Results

## Construction of B&J benchmark

The B&J benchmark is a novel benchmark designed to evaluate the clinical capabilities of AI models in musculoskeletal disorders. It contains 1,245 questions covering 42 common musculoskeletal disorders, with a distribution across seven anatomical sites: knee (26.5%), hip (20.5%), shoulder (15.9%), ankle (14.4%), elbow (12.0%), wrist (5.8%), and spine (4.9%).

We designed seven tasks mirroring clinician reasoning process (Table 2): knowledge recall (n=129); text interpretation (n=101); image interpretation (n=203); diagnosis generation (n=203); treatment planning (n=203); diagnosis and treatment reasoning (n=203). The details of the above tasks are shown in Table 1. Especially, diagnosis generation and treatment planning are multimodal tasks that necessitate the integrated analysis of a patient's medical history, physical examination findings, and radiological images to derive a final decision and its supporting rationale. The ground truth for the entire benchmark was meticulously developed by seven subspecialty experts, who formulated definite diagnostic/therapeutic decisions and the corresponding evidence-based reasoning pathways.

Furthermore, we designed an additional task to identify the core challenges in multimodal information processing by VLMs. By pairing clinical texts with anatomically mismatched images, this task evaluates whether VLMs can recognize the inconsistency of graphics and text during reasoning, thereby testing true multimodal integration, as opposed to generating responses based on modality-specific hallucinations.

Table 2. Overview of B&J benchmark evaluation scenarios.

| Clinical task | Model's capacity | Modality | Task format | Task description | Ground Truth | Data source | Sizes | Evaluation/ Metrics |
|---|---|---|---|---|---|---|---|---|
| Knowledge Recall | I: Medical Knowledge Recall | Text | MCQ (5 options) | Concept/theory questions about diagnosis & treatment | Correct answer choice | National Medical Licensing Examination; Orthopedic Attending Physician Qualification Examination | 129 | Automatic/ Accuracy |
| Unimodal Analysis | II: Clinical Note Interpretation | Text | MCQ (5 options) | Predict diagnosis/treatment from summarized clinical note | Correct answer choice | National Medical Licensing Examination; Orthopedic Attending Physician Qualification Examination | 101 | Automatic/ Accuracy |
| | III: Radiological Image Interpretation | Image | MCQ (5 options) | Predict diagnosis from X-ray/MRI/CT image | Correct answer choice | Real clinical cases from 8 centers | 203 | Automatic/ Accuracy |
| Multimodal Reasoning[a] | IV: Diagnoses Generation | Text & Image | Open-ended question | Provide complete diagnoses using full multimodal information | Diagnosis list from experts | Real clinical cases from 8 centers | 203 | GPT4-assisted & manual/Accuracy |
| | V: Diagnostic Reasoning | Text & Image | Open-ended question | Output chain-of-thought for diagnostic decisions | Reasoning chain of experts | Real clinical cases from 8 centers | 203 | Manual/Accuracy |
| | VI: Treatment planning | Text & Image | Open-ended question | Recommend therapies using full multimodal | Treatment plan from experts | Real clinical cases from 8 centers | 203 | Human GPT4-assisted |
| | | | | information | | | | |
| | VII: Treatment Reasoning | Text & Image | Open-ended question | Output chain-of-thought for treatment planning | Reasoning chain of experts | Real clinical cases from 8 centers | 203 | GPT4-assisted & manual/Accuracy |
| Additional Task | VIII: Text-Image Inconsistency Identification | Text & Image | Open-ended question | Recognize mismatched clinical text & image during reasoning | Correct answer choice | Real clinical cases from 8 centers with images replaced by unrelated anatomy site | 203 | Manual/Accuracy |

a: The multimodal information included patient's medical history, physical examination findings, and radiological images. To address the defect that LLMs cannot input image modalities, we replaced the radiological images with the corresponding radiology reports when evaluating LLMs.

**Performance Comparison among Different Models**

Our benchmark-based evaluation of eleven VLMs and six LLMs revealed a significant disparity in their performance on text-based versus image-based tasks (Fig. 3). In the MCQs requiring medical knowledge recall (Task I) and clinical note interpretation (Task II), SOTA VLMs such as GPT-4o achieved accuracy of 86.8% and 95.0%, respectively. In stark contrast, their performance on radiological image interpretation MCQs (Task III) was significantly poor, with accuracy ranging from 20.0% to 40.0%, showing only a marginal improvement over the 20% baseline for random guessing. We also observed a significant "text shortcut" tendency in VLMs when handling image-based tasks. This was evidenced by their propensity to generate diagnoses based on linguistic cues in the question or prompts, rather than performing a genuine analysis of the visual information within the images.

We also found that both VLMs and LLMs performed significantly better when tasks were presented in a structured MCQ format. For instance, SOTA LLM, like MedGPT, reached an accuracy of 96.0% in the clinical note interpretation task (Task II). However, a marked decline in performance was observed across all models when the tasks shifted to open-ended questions that demanded multimodal information integration and advanced clinical reasoning, such as diagnosis generation and treatment planning. On these more complex tasks, even the best-performing VLMs, including GPT-4o and Claude 3.5 Sonnet, failed to achieve an accuracy of 60.0%. When provided with radiology reports from clinicians, LLMs demonstrated improved accuracy in diagnosis generation, with MedGPT achieving 83.3% and DeepSeek-R1 reaching 70.0%. A similar improvement was noted in treatment planning, where DeepSeek-R1 achieved 53.30% accuracy. Nevertheless, there is still a huge gap between the models' reasoning processes and the chain-of-thought (CoT) demonstrated by human clinicians.

An unexpected finding was that VLMs specifically fine-tuned with medical data (e.g., LLaVA-Med and Med-Flamingo) did not demonstrate their superiority to general-purpose VLMs across all tasks. Their performance was particularly deficient in the open-ended questions (Task IV-VII), where accuracies were exceptionally low, ranging from 0% to 15%. Similarly, among the LLMs, the advantages of medical fine-tuning were not significant. The general-purpose model Deepseek-R1 exhibited performance comparable to that of the medical fine-tuned MedGPT.

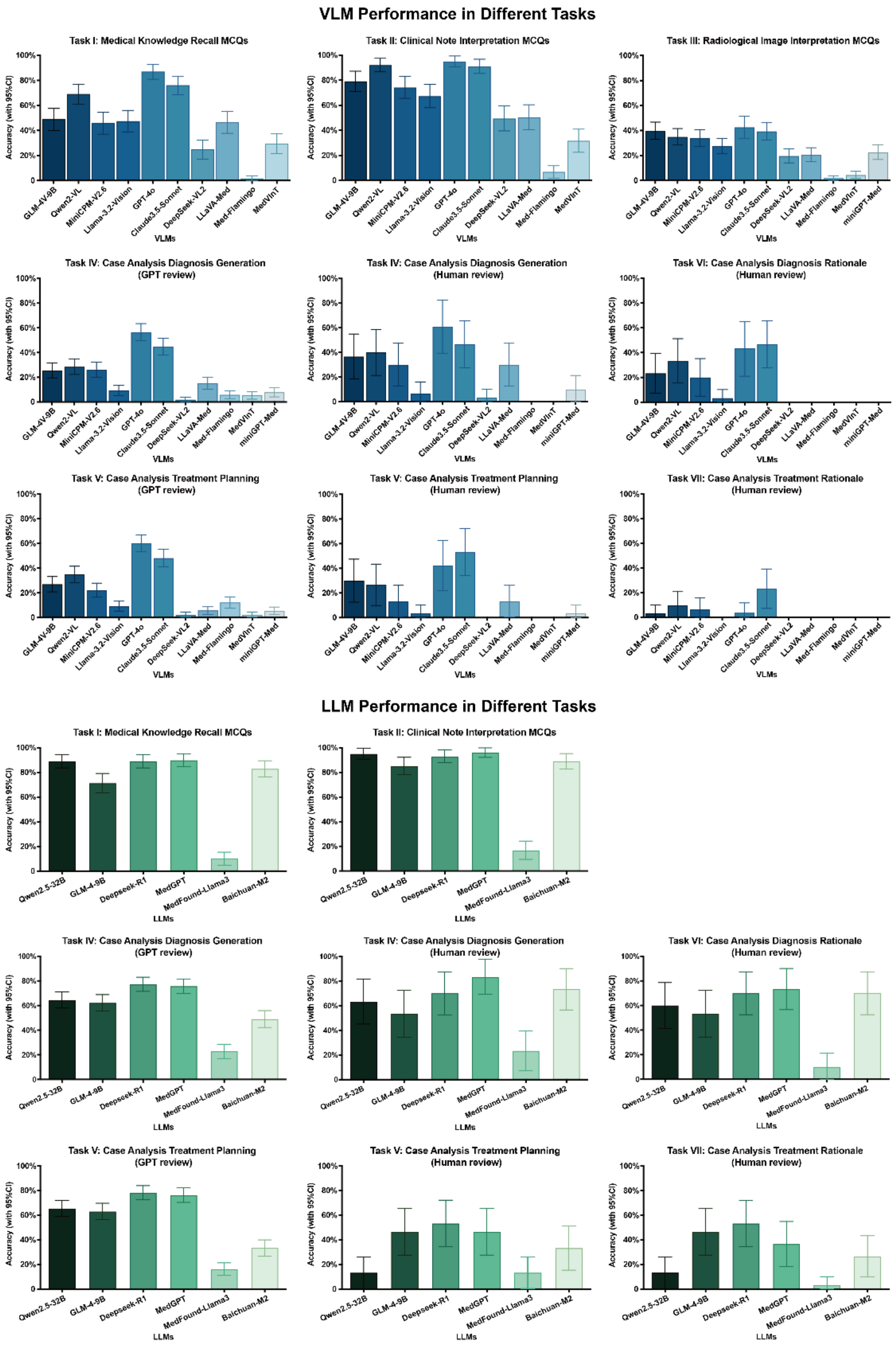

Figure 3. Performance of VLMs and LLMs in different tasks. It should be noted that the LLMs utilized radiology reports from clinicians in the task IV-VII, which means the results are not purely indicating these models' ability.

### Visual Hallucinations

Our preceding evaluation of radiological image interpretation MCQs (Task III) revealed significant limitations in the capabilities of VLMs. This observation raised a question: do these models truly integrate multimodal information for tasks such as diagnostic generation and treatment planning? To figure out this, we designed an additional experiment (Task VIII) to assess the VLMs' ability to identify inconsistency between clinical text and mismatched medical images during reasoning. The results indicate all the evaluation models failed in this task (Fig. 4). The highest accuracy was achieved by Claude 3.5, which detected only 35.0% of the mismatches. The performance of other models was considerably lower, with Llama-3.2-Vision at 15.3%, GPT-4o at 8.8%, and Qwen2-VL at a mere 1.0%. This poor performance suggests that the proclaimed "multimodal" capabilities of these VLMs often default to a text-dominant strategy. In this mode, the image does not serve as critical diagnostic evidence but instead acts as a primer for generating text-driven hallucinations, which are often erroneous.

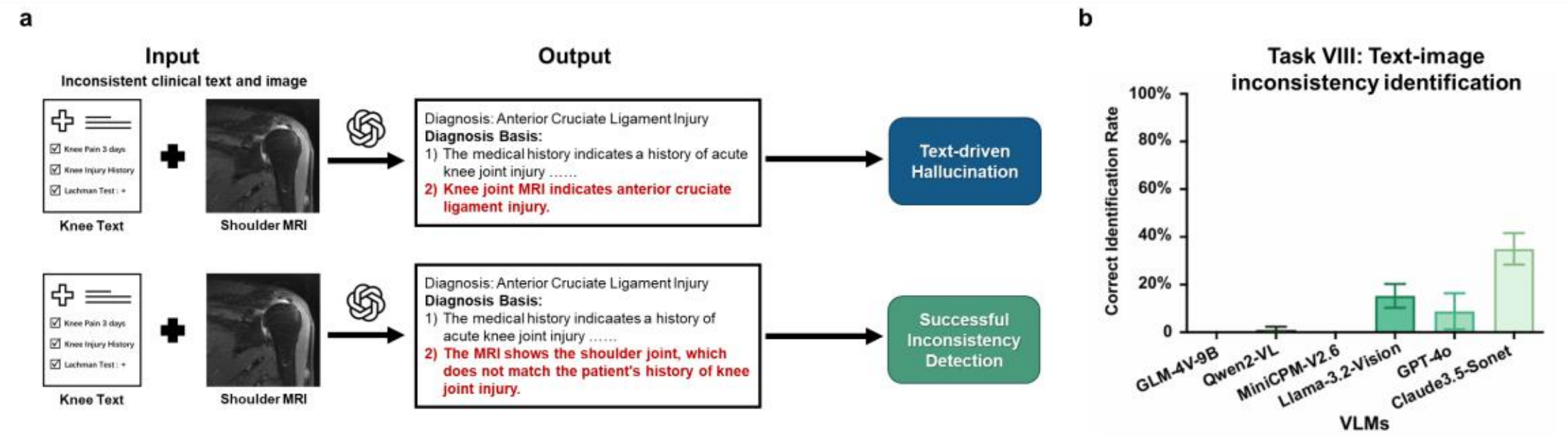

Figure 4. Visual Hallucinations Test. a) Examples of visual hallucinations. b) Performance of VLMs on Task VIII.

# Discussion

This study introduces the B&J benchmark, a structured evaluation framework comprising seven progressive tasks based on real-world clinical cases that replicate the complete clinical reasoning pathway from knowledge acquisition to treatment planning. By systematically assessing both general-purpose and medically specialized VLMs and LLMs, our benchmark redirects evaluation emphasis from factual knowledge recall to integrated clinical reasoning. The findings reveal clinically significant limitations in current AI systems, including constrained multimodal reasoning capacity, deficient radiological image interpretation, and prevalent text-driven hallucinations. These identified limitations highlight critical areas requiring methodological improvements before reliable clinical deployment can be achieved.

**The Pseudo-Reasoning Phenomenon and Pattern Recognition.** The evaluation results show that there is a significant disconnection between the text-based knowledge retrieval and clinical reasoning capabilities of current VLMs and LLMs. For instance, GPT-4o achieves 86.8% accuracy on text-based MCQs in Task I but scores only 3.9% on open-ended reasoning Task VII. This performance starkly contrasts with that of human clinicians, where a high scorer on a medical examination will not be unable to complete comprehensive case analysis. A detailed examination of model outputs suggests that what may appear to be reasoning is, in fact, a form of sophisticated pattern matching, a phenomenon we term "pseudo-reasoning". This observation aligns with emerging evidence indicating that model performance declines significantly when task complexity exceeds a certain threshold[21, 22].

Although these models can be optimized to function as powerful medical knowledge retrieval tools, their internal representations often lack the integration required for genuine clinical reasoning, leading them to depend on statistical regularities from training data[23]. This limitation also raises concerns about potential data contamination in publicly available benchmarks[24]. Consequently, VLMs and LLMs should be positioned as clinical copilots rather than primary decision-makers. They can generate differential diagnoses and treatment options, providing clinicians with data-driven possibilities and supporting evidence. However, final decisions must integrate the clinician's causal knowledge, clinical experience, and patient-specific factors to verify, reject, or optimize AI-generated content. The common disclaimer that "model outputs are for reference only and should be used under professional guidance" reflects this essential reality, though this crucial disclaimer is frequently overlooked by the public, who may not recognize the limitations of AI systems in medical contexts.

**Systematic Deficiencies in Medical Image Interpretation.** Our study reveals

that current VLMs generally lack effective visual interpretation capabilities for medical images. The results indicate that both general-purpose and medical-specific VLMs exhibit severe deficiencies in image recognition. Specifically, while the accuracy of text-based MCQs (Task I and II) can reach over 90%, the accuracy of image recognition (Task III) drops sharply to 30%, within the expected range for random answering. Interestingly, it is worth noting that medical-specific VLMs demonstrate no advantage in this field, and their performance even lags significantly behind leading general-purpose models such as GPT-4o, indicating that training in specialized domains is insufficient to compensate for the inherent limitations of the basic visual encoders. This phenomenon stems from a dual challenge: on the one hand, high-quality orthopedic image datasets with expert annotations are scarce[25, 26]; On the other hand, the results show that the benefits brought by merely fine-tuning the weak foundation model with specific data are negligible and may even have negative impacts[27, 28]. Clinical image analysis in the real world is far more complex than multiple-choice questions. Clinicians not only need to classify the diagnosis from images but also formulate treatment plans by carefully observing the granularity details (such as the degree of fracture displacement, and ligament integrity). Current models lack such fine-grained visual interpretation capability, making it difficult for them to handle vision-driven clinical decision-making[29].

**Failure of Multimodal Fusion and the Emergence of Text-Driven Hallucinations.** While some studies have suggested that VLMs can successfully integrate visual and textual information to form coherent diagnostic reasoning[30], sometimes even reportedly surpassing human clinicians[31, 32], our findings present a contrasting reality. Through controlled visual hallucination tests on leading VLMs, we observed that these models consistently default to textual reasoning when confronted with conflicting text-image pairs, often disregarding critical visual evidence. This behavior underscores a fundamental weakness in their multimodal fusion mechanisms, revealing that their supposed integrative capabilities are not yet reliably established.

This systematic bias toward text stems from both architectural and optimization constraints. Architecturally, the prevalent cross-attention mechanism treats text tokens as active "queries" while reducing visual features to passive "keys/values," creating a fundamental modality imbalance that marginalizes image evidence, especially during conflict[33, 34]. From an optimization perspective, training objectives such as next-token prediction reward linguistically plausible narratives, and subsequent alignment through human feedback further amplifies the preference for text-based outputs. In effect, these limitations reduce the VLM to little more than a language model with a visual suggestion box, rather than a tool capable of genuine multimodal reasoning. This failure poses a direct clinical danger: models may hallucinate findings in normal images to match the text, causing over-diagnosis and overtreatment; or they may dismiss true image abnormalities amid sparse text, resulting in missed diagnoses and delayed care. Such systematic errors risk converting AI from a supportive aid into a significant source of clinical mistakes[35-37].

According to the current deficiencies of VLMs and LLMs mentioned above, the answer to the question of "can contemporary AI models truly achieve clinical competence" is no. Instead, its integration into medicine will likely follow a phased, task-specific pathway. The most immediate and feasible applications lie in text-based domains, such as knowledge retrieval[2] and clinical summarization[38], which leverage the models' strengths in natural language processing while avoiding their underdeveloped multimodal reasoning. Advancing to the core clinical tasks of image interpretation and diagnostic-therapeutic planning will require not only vast, meticulously curated medical image datasets but also fundamental breakthroughs in underlying algorithms.

While this study presents a clinically grounded multimodal benchmark, we still recognize its limitations. Firstly, our benchmark is focused exclusively on musculoskeletal disorders. Consequently, the specific performance of models may vary when applied to other medical disciplines with different data modality preference and treatment goal, such as dermatology or oncology. However, the fundamental shortcomings we identified, including constrained multimodal reasoning, visual hallucination, and a disconnect between knowledge retrieval and clinical reasoning, are likely to be representative of broader challenges facing current VLMs and LLMs.

Secondly, due to the technical constraints of current VLMs, our benchmark is reliance on 2D static images, which inherently simplifies volumetric data like MRI, leading to a loss of critical spatial information[39, 40]. Thirdly, our tasks focus on “textbook-like” clinical cases, where signs and symptoms are highly consistent, do not fully capture the complexity of real-world cases, which often involve atypical presentations or contradictory evidence. This simplification may lead to an overestimation of model performance. Therefore, future iterations of the benchmark should aim to incorporate full 3D volumetric data as model architecture evolve and expand the dataset to include a collection of challenging and atypical cases to more rigorously assess model robustness and clinical capability[41].

# Conclusion

In conclusion, the B&J benchmark provides a solid foundation for evaluating and promoting the application of VLMs and LLMs in real clinical scenarios. We believe that through continuous iteration and improvement of this evaluation benchmark, it will effectively promote the development of safer and more reliable medical AI models.

# References

1. Singhal K, Tu T, Gottweis J, Sayres R, Wulczyn E, Amin M, et al. Toward expert-level medical question answering with large language models. NAT MED 2025;31:943-50. DOI:10.1038/s41591-024-03423-7.
2. Singhal K, Azizi S, Tu T, Mahdavi SS, Wei J, Chung HW, et al. Large language models encode clinical knowledge. NATURE 2023;620:172-80. DOI:10.1038/s41586-023-06291-2.
3. Zeng D, Qin Y, Sheng B, Wong TY. DeepSeek's "Low-Cost" Adoption Across China's Hospital Systems: Too Fast, Too Soon? JAMA-J AM MED ASSOC 2025;333:1866-9. DOI:10.1001/jama.2025.6571.
4. Sandmann S, Hegselmann S, Fujarski M, Bickmann L, Wild B, Eils R, et al. Benchmark evaluation of DeepSeek large language models in clinical decision-making. NAT MED 2025;31:2546-9. DOI:10.1038/s41591-025-03727-2.

5. Panteli D, Adib K, Buttigieg S, Goiana-da-Silva F, Ladewig K, Azzopardi-Muscat N, et al. Artificial intelligence in public health: promises, challenges, and an agenda for policy makers and public health institutions. LANCET PUBLIC HEALTH 2025;10:e428-32. DOI:10.1016/S2468-2667(25)00036-2.
6. Ayers JW, Poliak A, Dredze M, Leas EC, Zhu Z, Kelley JB, et al. Comparing Physician and Artificial Intelligence Chatbot Responses to Patient Questions Posted to a Public Social Media Forum. JAMA INTERN MED 2023;183:589-96. DOI:10.1001/jamainternmed.2023.1838.
7. Pal A, Umapathi LK, Sankarasubbu M. MedMCQA: A Large-scale Multi-Subject Multi-Choice Dataset for Medical domain Question Answering. In: Proceedings of the Conference on Health, Inference, and Learning; Proceedings of Machine Learning Research: PMLR; 2022. p. 248-60.
8. Jin D, Pan E, Oufattole N, Weng W, Fang H, Szolovits P. What Disease does this Patient Have? A Large-scale Open Domain Question Answering Dataset from Medical Exams. September 28, 2020 (https://arxiv.org/abs/2009.13081). Preprint.
9. Jin Q, Dhingra B, Liu Z, Cohen WW, Lu X. PubMedQA: A Dataset for Biomedical Research Question Answering. September 13, 2019 (https://arxiv.org/abs/1909.06146). Preprint.
10. Rodman A, Zwaan L, Olson A, Manrai AK. When It Comes to Benchmarks, Humans Are the Only Way. NEJM AI 2025;2:AIe2500143. DOI:10.1056/AIe2500143.
11. Griot M, Vanderdonckt J, Yuksel D, Hemptinne C. Pattern Recognition or Medical Knowledge? The Problem with Multiple-Choice Questions in Medicine. In: Proceedings of the 63rd Annual Meeting of the Association for Computational Linguistics (Volume 1: Long Papers); Vienna, Austria: Association for Computational Linguistics; 2025. p. 5321-41.
12. Templin T, Fort S, Padmanabham P, Seshadri P, Rimal R, Oliva J, et al. Framework for bias evaluation in large language models in healthcare settings. NPJ DIGIT MED 2025;8:414. DOI:10.1038/s41746-025-01786-w.
13. van de Sande D, Chung EFF, Oosterhoff J, van Bommel J, Gommers D, van Genderen ME. To warrant clinical adoption AI models require a multi-faceted

implementation evaluation. NPJ DIGIT MED 2024;7:58. DOI:10.1038/s41746-024-01064-1.

14. Global, regional, and national burden of neck pain, 1990-2020, and projections to 2050: a systematic analysis of the Global Burden of Disease Study 2021. LANCET RHEUMATOL 2024;6:e142-55. DOI:10.1016/S2665-9913(23)00321-1.

15. Global, regional, and national burden of osteoarthritis, 1990-2020 and projections to 2050: a systematic analysis for the Global Burden of Disease Study 2021. LANCET RHEUMATOL 2023;5:e508-22. DOI:10.1016/S2665-9913(23)00163-7.

16. Global, regional, and national burden of other musculoskeletal disorders, 1990-2020, and projections to 2050: a systematic analysis of the Global Burden of Disease Study 2021. LANCET RHEUMATOL 2023;5:e670-82. DOI:10.1016/S2665-9913(23)00232-1.

17. Hartvigsen J, Hancock MJ, Kongsted A, Louw Q, Ferreira ML, Genevay S, et al. What low back pain is and why we need to pay attention. LANCET 2018;391:2356-67. DOI:10.1016/S0140-6736(18)30480-X.

18. Lin I, Wiles L, Waller R, Goucke R, Nagree Y, Gibberd M, et al. What does best practice care for musculoskeletal pain look like? Eleven consistent recommendations from high-quality clinical practice guidelines: systematic review. BRIT J SPORT MED 2020;54:79-86. DOI:10.1136/bjsports-2018-099878.

19. Nori H, King N, McKinney SM, Carignan D, Horvitz E. Capabilities of GPT-4 on Medical Challenge Problems. March 20, 2023 (https://arxiv.org/abs/2303.13375). Preprint.

20. Patterson D, Gonzalez J, Urs Hölzle QL, Liang C, Munguia L, Rothchild D, et al. The Carbon Footprint of Machine Learning Training Will Plateau, Then Shrink. COMPUTER 2022;55:18-28. DOI:10.1109/MC.2022.3148714.

21. Shojaee P, Mirzadeh I, Alizadeh-Vahid K, Horton M, Bengio S, Farajtabar M. The Illusion of Thinking: Understanding the Strengths and Limitations of Reasoning Models via the Lens of Problem Complexity. November 20, 2025 (https://arxiv.org/abs/2506.06941). Preprint.

22. Mirzadeh I, Alizadeh K, Shahrokhi H, Tuzel O, Bengio S, Farajtabar M. GSM-

Symbolic: Understanding the Limitations of Mathematical Reasoning in Large Language Models. August 27, 2025 (https://arxiv.org/abs/2410.05229). Preprint.

23. Bedi S, Jiang Y, Chung P, Koyejo S, Shah N. Fidelity of Medical Reasoning in Large Language Models. JAMA NETW OPEN 2025;8:e2526021. DOI:10.1001/jamanetworkopen.2025.26021.

24. Deng C, Zhao Y, Tang X, Gerstein MB, Cohan A. Investigating Data Contamination in Modern Benchmarks for Large Language Models. In: Proceedings of the 2024 Conference of the North American Chapter of the Association for Computational Linguistics: Human Language Technologies (Volume 1: Long Papers); Mexico City, Mexico: Association for Computational Linguistics; 2024. p. 8706-19.

25. Schafer R, Nicke T, Hofener H, Lange A, Merhof D, Feuerhake F, et al. Overcoming data scarcity in biomedical imaging with a foundational multi-task model. Nat Comput Sci 2024;4:495-509. DOI:10.1038/s43588-024-00662-z.

26. Wang S, Li C, Wang R, Liu Z, Wang M, Tan H, et al. Annotation-efficient deep learning for automatic medical image segmentation. NAT COMMUN 2021;12:5915. DOI:10.1038/s41467-021-26216-9.

27. Jeong DP, Mani P, Garg S, Lipton ZC, Oberst M. The Limited Impact of Medical Adaptation of Large Language and Vision-Language Models. June 28, 2025 (https://arxiv.org/abs/2411.08870). Preprint.

28. Jeong DP, Garg S, Lipton ZC, Oberst M. Medical Adaptation of Large Language and Vision-Language Models: Are We Making Progress? In: Proceedings of the 2024 Conference on Empirical Methods in Natural Language Processing; Miami, Florida, USA: Association for Computational Linguistics; 2024. p. 12143–12170.

29. Liang X, Li X, Li F, Jiang J, Dong Q, Wang W, et al. MedFILIP: Medical Fine-Grained Language-Image Pre-Training. IEEE J BIOMED HEALTH 2025;29:3587-97. DOI:10.1109/JBHI.2025.3528196.

30. Wang S, Hu M, Li Q, Safari M, Yang X. Capabilities of GPT-5 on Multimodal Medical Reasoning. August 13, 2025 (https://arxiv.org/abs/2508.08224). Preprint.

31. Goh E, Gallo R, Hom J, Strong E, Weng Y, Kerman H, et al. Large Language Model Influence on Diagnostic Reasoning: A Randomized Clinical Trial. JAMA NETW

OPEN 2024;7:e2440969. DOI:10.1001/jamanetworkopen.2024.40969.
32. Kaczmarczyk R, Wilhelm TI, Martin R, Roos J. Evaluating multimodal AI in medical diagnostics. NPJ DIGIT MED 2024;7:205. DOI:10.1038/s41746-024-01208-3.
33. Wu H, Tang M, Zheng X, Jiang H. When Language Overrules: Revealing Text Dominance in Multimodal Large Language Models. August 14, 2025 (https://arxiv.org/abs/2508.10552). Preprint.
34. Zheng X, Wu H, Wang X, Jiang H. Unveiling Intrinsic Text Bias in Multimodal Large Language Models through Attention Key-Space Analysis. October 30, 2025 (https://arxiv.org/abs/2510.26721). Preprint.
35. Chen J, Yang D, Wu T, Jiang Y, Hou X, Li M, et al. Detecting and Evaluating Medical Hallucinations in Large Vision Language Models. June 14, 2024 (https://arxiv.org/abs/2406.10185). Preprint.
36. Thirunavukarasu AJ, Ting DSJ, Elangovan K, Gutierrez L, Tan TF, Ting DSW. Large language models in medicine. NAT MED 2023;29:1930-40. DOI:10.1038/s41591-023-02448-8.
37. Zhu Z, Zhang Y, Zhuang X, Zhang F, Wan Z, Chen Y, et al. Can We Trust AI Doctors? A Survey of Medical Hallucination in Large Language and Large Vision-Language Models. In: Findings of the Association for Computational Linguistics: ACL 2025; Vienna, Austria: Association for Computational Linguistics; 2025. p. 6748-69.
38. Van Veen D, Van Uden C, Blankemeier L, Delbrouck J, Aali A, Bluethgen C, et al. Adapted large language models can outperform medical experts in clinical text summarization. NAT MED 2024;30:1134-42. DOI:10.1038/s41591-024-02855-5.
39. Meyer A, Chlebus G, Rak M, Schindele D, Schostak M, van Ginneken B, et al. Anisotropic 3D Multi-Stream CNN for Accurate Prostate Segmentation from Multi-Planar MRI. COMPUT METH PROG BIO 2021;200:105821. DOI:10.1016/j.cmpb.2020.105821.
40. Gonzalez Ballester MA, Zisserman AP, Brady M. Estimation of the partial volume effect in MRI. MED IMAGE ANAL 2002;6:389-405. DOI:10.1016/s1361-8415(02)00061-0.

41. Tang Y, Yang D, Li W, Roth HR, Landman BA, Xu D, et al. Self-Supervised Pre-Training of Swin Transformers for 3D Medical Image Analysis. In: 2022 IEEE/CVF Conference on Computer Vision and Pattern Recognition (CVPR); New Orleans, LA, USA: IEEE; 2022. p. 20698-708.

# Disclosures

This work was funded by the National Natural Science Foundation of China (Grant number 82441025), Beijing Municipal Natural Science Foundation (Grand number L242104), and Peking University Third Hospital Cross-Disciplinary Joint Special Fund (Grant number BYSYJC2024001). We would like to express our gratitude to Weili Fu, Bingchuan Liu, Xiaodong Ju, Fan Yang, Zhenxing Shao, Dan Xiao, and Jian Huang for their efforts in constructing the benchmark and ground truth. We also thank Haodong Bai for his assistance in model deployment.